\documentclass[journal,twoside,web]{ieeecolor}
\usepackage{tmi}
\usepackage{cite}
\usepackage{amsmath,amssymb,amsfonts}
\usepackage{algorithmic}
\usepackage{graphicx}
\usepackage{textcomp, makecell}
\usepackage{hyperref, multirow}
\usepackage[nameinlink,capitalize]{cleveref}
\def\equationautorefname~#1\null{Equation (#1)\null}

\def\BibTeX{{\rm B\kern-.05em{\sc i\kern-.025em b}\kern-.08em
    T\kern-.1667em\lower.7ex\hbox{E}\kern-.125emX}}
\begin{document}
\title{Echo-SyncNet: Self-supervised Cardiac View Synchronization in Echocardiography}
\author{Fatemeh~Taheri~Dezaki,~Christina~Luong,~Tom~Ginsberg,~Robert~Rohling,~Ken~Gin, Purang~Abolmaesumi,~and~Teresa~Tsang

\thanks{F. Taheri Dezaki is with the Department of Electrical and Computer Engineering, The University of British Columbia, Vancouver, BC V6T 1Z4, Canada (e-mail:fatemeht@ece.ubc.ca).}%
\thanks{C. Luong and K. Gin are with Vancouver General Hospital Echocardiography Laboratory, Division of Cardiology, Department of Medicine, The University of British Columbia, Vancouver, BC V5Z 1M9, Canada.(e-mail: christina.luong@ubc.ca, Kenneth.gin@vch.ca)}%
\thanks{T. Ginsberg is with the Department of Engineering Physics, The University of British Columbia, Vancouver, BC V6T 1Z4, Canada (e-mail:tom.ginsberg@alumni.ubc.ca).}%

\thanks{R. Rohling is with the Department of Electrical and Computer Engineering and the Department of Mechanical Engineering, The University of British Columbia, Vancouver, BC V6T 1Z4, Canada (e-mail: rohling@ece.ubc.ca,)}%

\thanks{P. Abolmaesumi is Co-Principal Investigator for the CIHR-NSERC grant supporting this work and is with the Department of Electrical and Computer Engineering, The University of British Columbia, Vancouver, BC V6T 1Z4, Canada. (e-mail: purang@ece.ubc.ca)}%
\thanks{T. Tsang is Principal Investigator of the CIHR-NSERC grant supporting this work and is the Director of the Vancouver General Hospital and University of British Columbia Echocardiography Laboratories, Division of Cardiology, Department of Medicine, The University of British Columbia, Vancouver, BC V5Z 1M9, Canada. (e-mail:t.tsang@ubc.ca).}%

\thanks{F. Taheri Dezaki and C. Luong 
have contributed equally to this work.}%
\thanks{P. Abolmaesumi and T. Tsang have contributed equally to this work.}%

}

\maketitle

\begin{abstract}
In echocardiography (echo), an electrocardiogram (ECG) is conventionally used to temporally align different cardiac views for assessing critical measurements. However, in emergencies or point-of-care situations, acquiring an ECG is often not an option, hence motivating the need for alternative temporal synchronization methods. Here, we propose Echo-SyncNet, a self-supervised learning framework to synchronize various cross-sectional 2D echo series without any human supervision or external inputs. The proposed framework takes advantage of two types of supervisory signals derived from the input data: spatiotemporal patterns found between the frames of a single cine (intra-view self-supervision) and interdependencies between multiple cines (inter-view self-supervision). The combined supervisory signals are used to learn a feature-rich and low dimensional embedding space where multiple echo cines can be temporally synchronized. Two intra-view self-supervisions are used, the first is based on the information encoded by the temporal ordering of a cine (temporal intra-view) and the second on the spatial similarities between nearby frames (spatial intra-view). The inter-view self-supervision is used to promote the learning of similar embeddings for frames captured from the same cardiac phase in different echo views. We evaluate the framework with multiple experiments: 1) Using data from 998 patients, Echo-SyncNet shows promising results for synchronizing Apical 2 chamber and Apical 4 chamber cardiac views, which are acquired spatially perpendicular to each other; 2) Using data from 3070 patients, our experiments reveal that the learned representations of Echo-SyncNet outperform a supervised deep learning method that is optimized for automatic detection of fine-grained cardiac cycle phase; 3) We go one step further and show the usefulness of the learned representations in a one-shot learning scenario of cardiac key-frame detection. Without any fine-tuning, key frames in 1188 validation patient studies are identified by synchronizing them with only one labeled reference cine. We do not make any prior assumption about what specific cardiac views are used for training, and hence we show that Echo-SyncNet can accurately generalize to views not present in its training set. Project repository:  \href{{github.com/fatemehtd/Echo-SyncNet}}{\texttt{github.com/fatemehtd/Echo-SyncNet}}. 
\end{abstract}


\section{Introduction}
\label{sec:introduction}
\IEEEPARstart{E}{chocardiography} (echo) plays an important role in cardiac imaging and provides a non-invasive, low-cost, and widely available diagnostic tool for the comprehensive evaluation of cardiac structure and function. A cine is an ultrasound video that captures a 2D cross-section of the heart from a specific orientation. Several common cine views, characterized by the plane in which they are captured, are frequently taken by cardiac specialists and used for qualitative and/or quantitative analysis of cardiac function.

To evaluate important cardiac measurements such as left ventricular ejection fraction (EF), detection of wall motion abnormality (WMA) or severity of aortic stenosis (AS) clinicians often use several cines from various views. When analysing multiple views, temporal synchronization or simply cine synchronization, becomes critical for making sure the cardiac phase is the same at every instant in all views. Cine synchronization is a challenging task considering the large quality variations in typical echo and the lack of obvious structural correspondences between views. Normally, synchronization of multiple cardiac views is based on concurrently recorded electrocardiogram (ECG) signals. ECGs can be easily synchronized allowing clinicians to view multiple cines play together with near-perfect temporal correspondance~\cite{lang2015recommendations}. Cine synchronization allows for better visualization of the cardiac structure, more accurate measurements and hence, a simpler and more reliable cardiac assessment.

In machine learning literature, several research groups have taken advantage of synchronized echo views in developing their models~\cite{Delaram_EF2019,MultiView_ShouLiMICCAI2019,MultiView_ShouLiTMI2019}. In echo based EF prediction, Behnami et al.~\cite{Delaram_EF2019} showed that a prediction model performs better when two synchronized views, apical 4 chamber (AP4) and apical 2 chamber (AP2),  were fed to the model. 

In~\cite{MultiView_ShouLiMICCAI2019}, Li et al. applied their proposed segmentation method on synchronized sequences from three views, Apical 3 chamber (AP3), AP4, and AP2. To produce synchronized echo cines, it is required to have either expert annotations of the cardiac phase or the presence of an ECG. Unfortunately, expert annotations or ECG-based cine synchronization is often not feasible in emergencies or point-of-care~\cite{dezaki2018cardiac}. The use of ECG sensors can be also prohibited when there is sterilization barriers in clinics, such as those we are facing with the ongoing COVID-19 pandemic.

With the advent of deep learning in computer vision, the medical imaging community has rapidly adopted and developed algorithms to assist physicians in clinical routine. Nevertheless, the overall performance of fully-supervised learning highly depends on the size of the annotated data-set, resulting in sub-optimal performance for smaller data-sets. Generating expert annotations of patient data at scale is non-trivial, expensive, and time-consuming. Consequently, unsupervised and self-supervised representation learning is gaining attention to overcome the problem of learning under a low data regimen and to reduce the required cost for expert annotation. In self-supervised learning, supervision comes from the data itself. The only and most important step is to set the learning objectives properly. 

Numerous attempts have been made on self-supervised representation learning with non-medical images and videos. Self-supervised approaches can be generally categorized into encoder and encoder-decoder architectures. In the first category, a convolutional neural network encoder is trained based on developing pretext tasks that learn meaningful representations~\cite{Misra2016ShuffleAL,SS_odd_one_out_2017,SS_arrow_of_time_2018}. In the encoder-decoder architectures~\cite{SS_tracking_2015,SS_correspondence_2019,SS_colorizing_2018}, a model is trained based on reconstruction tasks. Self-supervised learning has been applied to many computer vision problems in image and video representation. To stay aligned with the work in this paper, we only review prior methods related to video learning. In~\cite{SS_tracking_2015,SS_correspondence_2019}, tracking moving objects in videos is used as a tool for learning visual representations.  Several other self-supervised tasks rely on learning representations based on the chronological order of frames~\cite{Misra2016ShuffleAL,SS_odd_one_out_2017,SS_arrow_of_time_2018}. Moreover, video colorization by copying colors from a reference frame to target frames is proposed~\cite{SS_colorizing_2018} as a means of self-supervised video representation learning. 

\begin{figure}
    \centering
    \includegraphics[width=\linewidth]{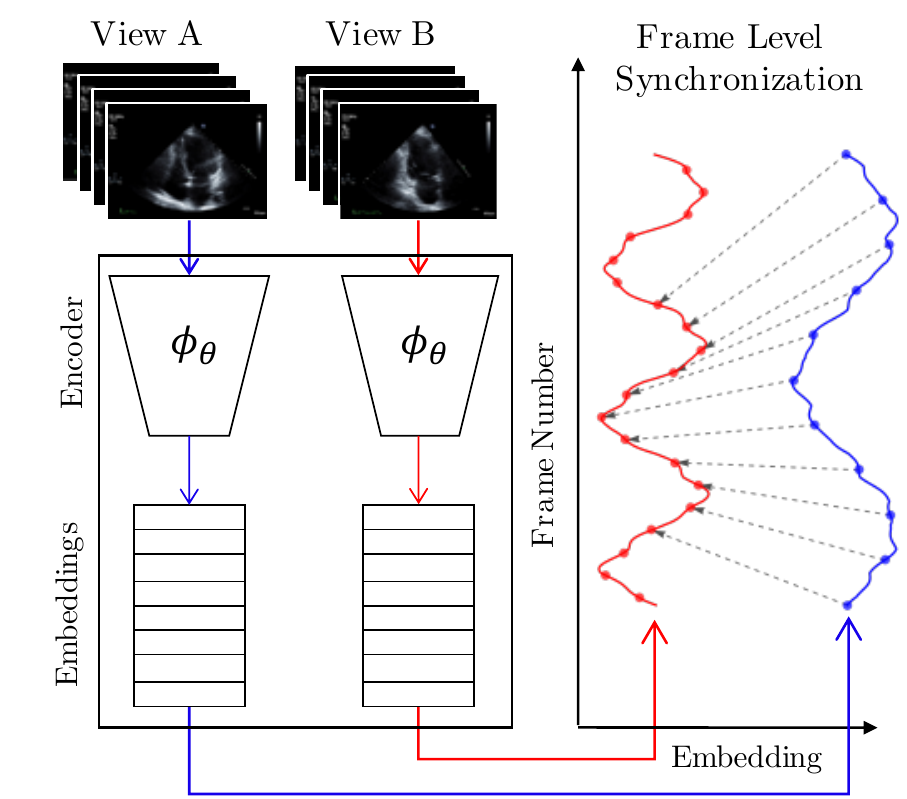}
    \caption{Overview of the proposed Echo-SyncNet framework. Two cardiac echo views are mapped to embeddings via a shared encoder $\phi_\theta$. The embeddings are designed to enable a robust temporal synchronization.}
    \label{fig:overview}
\end{figure}

Most of the prominent self-supervised methods have been derived in the context of natural images and videos, without considering the unique properties that medical imaging has to offer.
In echo, with the lack of ECG or other external annotations related to synchronization, we propose a self-supervised framework, Echo-SyncNet (\autoref{fig:overview}), to temporally synchronize cardiac echo cines acquired from different views. We hypothesize that consistent temporal and structural information across cardiac views can be leveraged to establish a robust learning objective. Echo-SyncNet exploits two types of supervisory signals derived from the input data: spatiotemporal patterns found between the frames of a single cine (intra-view self-supervision), and interdependencies between multiple cines (inter-view self-supervision). The combined supervisory signals are used to learn a feature-rich and low-dimensional embedding space, where multiple echo cines can be temporally synchronized. Using data from 998 patients, our proposed framework shows promising results for synchronizing AP2 and AP4 cardiac views, which are acquired spatially perpendicular to each other. Furthermore, using data from 3070 patients, we demonstrate that the learned representations of our proposed framework outperform a supervised deep learning method that is optimized for automatic detection of fine-grained cardiac phase-detection. We do not make any prior assumption about specific cardiac views used for training, and hence we show that Echo-SyncNet can accurately generalize to views not present in its training set. Moreover, in an experiment in a one-shot learning scenario using only one cine with labeled key frames, we show the usefulness of the learned embeddings in identifying key frames in unseen validation cines, significantly reducing the burden of manual annotation.
\smallbreak
In summary, our contributions are:
\begin{itemize}
    \item a self-supervised framework for fine-grained temporal alignment of cardiac echo cines captured from different echo views;
    \item evaluation of the proposed method on a large echocardiography data set, and demonstration of its effectiveness to synchronize various cardiac views;
    \item a thorough analysis of the use of our framework for the detection of fine-grained cardiac phases, specifically end-diastolic (ED) and end-systolic (ES) cardiac events, without the need for significant labeled data; and,
    \item showing the extension of our framework to the synchronization of multiple types of cardiac views---even views not present in the training set---with empirical evidence.
\end{itemize}
Additionally our source code is made available\footnote{\href{{github.com/fatemehtd/Echo-SyncNet}}{Code availability: \texttt{github.com/fatemehtd/Echo-SyncNet}}} included with an example usage of Echo-SyncNet on the Stanford EchoNet-Dynamic public dataset and the supplementary materials. 

\section{Methods}
In this section, we describe the proposed Echo-SyncNet approach and its components.
\subsection{Echo-SyncNet}
 In practice, 2D echo cines are obtained from different imaging windows, capturing various cross-sections of the heart. For example, AP4 and AP2 are two perpendicular cardiac views that are most commonly used to study the left ventricle and quantify its systolic function~\cite{lang2015recommendations}. 


We intend to learn an embedding space where multiple echo cines from different views can be synchronized temporally. Each cine by itself or jointly with other cines can contribute to the learning of a view point-invariant embedding space. By learning from multi-view cines, the generated representations effectively disentangle functional attributes such as contraction or expansion of the left ventricle, and opening or closing of the mitral valve. 

For a given cine, every frame $f_i$ with height, width, and channels $H$, $W$, and $C$, respectively, is encoded as $p_i = \phi_{\theta}(f_i)$, where $\phi_{\theta}: \mathbb{R}^{H\times W\times C}\mapsto \mathbb{R}^d$ is the encoder network parameterized by $\theta$ which maps video frames to $d$-dimensional feature vectors. Echo-SyncNet learns embeddings by a weighted sum of three losses as in (\ref{totalloss}): temporal and spatial intra-view and inter-view losses, where $\mathcal{L}_i$ represents a loss function and $\lambda_i$ is an associated weight:

\begin{align}
\begin{split}
\mathcal{L}_{\mathrm{Echo-SyncNet}} &=  \lambda_1 \mathcal{L}_{\mathrm{intra-spatial}}\\
&+\lambda_2 \mathcal{L}_{\mathrm{intra-temporal}}+\lambda_3 \mathcal{L}_{\mathrm{inter-view}} .
\end{split}
\label{totalloss}
\end{align}

 The loss functions that constitute the total loss leverage a shared encoder network architecture and are combined in a single multitask training framework outlined in \autoref{implementation}. Therefore, the combined losses backpropagate through the same network parameters. The weights for the relative contribution of each loss were discovered through a hyperparameter search. 

\begin{figure*}[ht]
\begin{center}
\includegraphics[width=\textwidth]{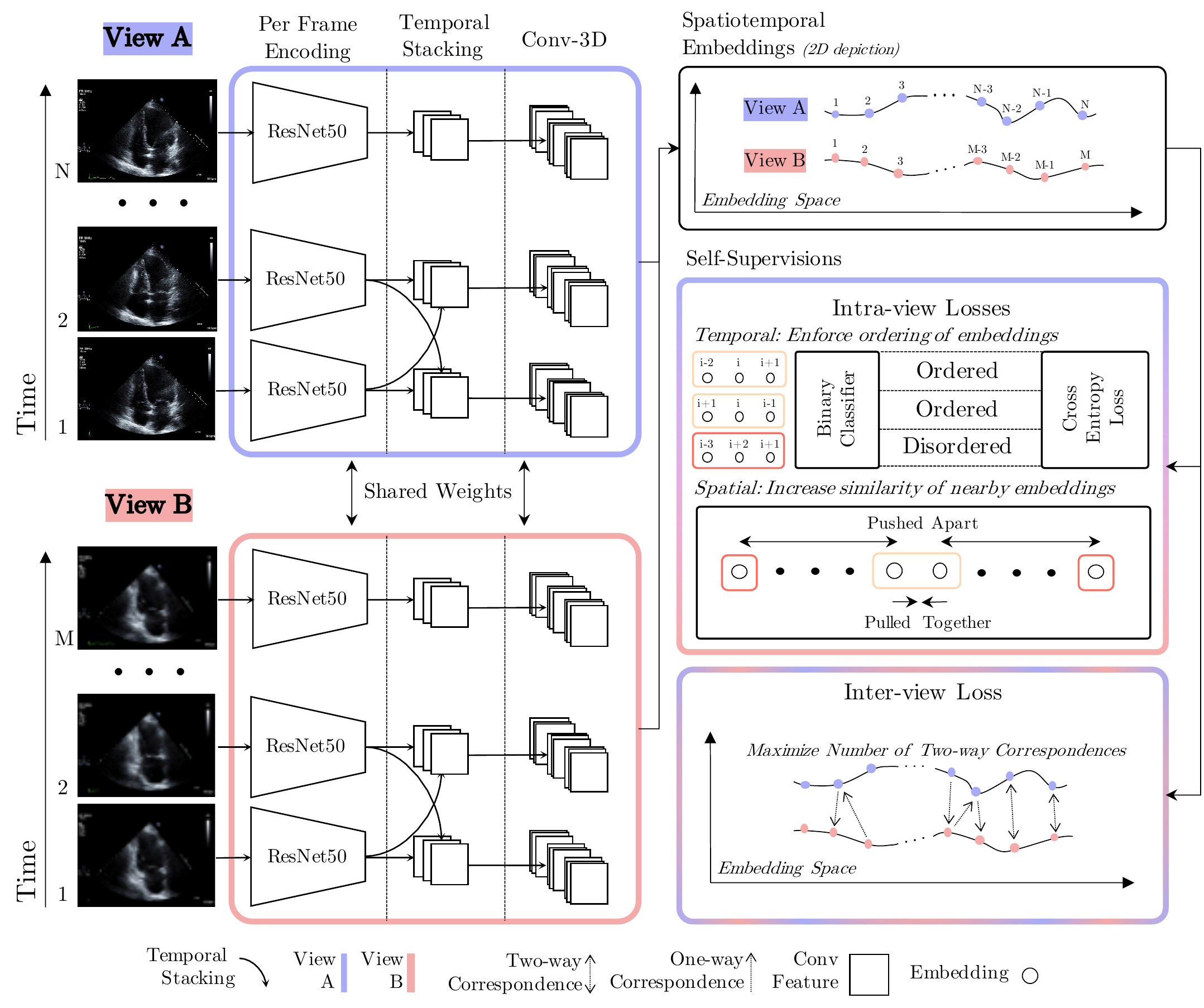}
\end{center}
\caption{The full training framework for Echo-SyncNet. The base 2D network gives an embedding for each frame after which $k$ context frames are temporally stacked and passed through the 3D convolutional network. Losses from three self-supervisions on the encodings are linearly combined and used to train the encoder network to produce embeddings optimized for temporal synchronization. The two intra-view supervisions are used to capture temporal and spatial features found within individual cardiac echos. The inter-view supervision (synchronization loss) is used to maximize the similarity of embedded features across different echo views. It should be noted that this framework may be be trained with any two echo views; here, we show the AP2 and AP4 views as an example.}
\label{fig:framework}
\end{figure*}

 A visual description of the entire process of Echo-SyncNet is shown in ~\autoref{fig:framework} and a detailed explanation of each component is provided in the following subsections.

\subsection{Temporal and Spatial Intra-view Self-supervision}
The goal of the Echo-SyncNet encoder, $\phi_\theta$, is to produce a latent representation of a cine that is fine-tuned for robust temporal alignment. To ensure that $\phi_\theta$ possess an adequate understanding of individual cines, we formulate two self-supervised learning objectives to take advantage of hidden spatial and temporal features.
\smallbreak
\subsubsection{Temporal Intra-view Self-supervision} 
\label{temp-intra}
\begin{figure}[t]
\begin{center}
\includegraphics[width=1.0\columnwidth]{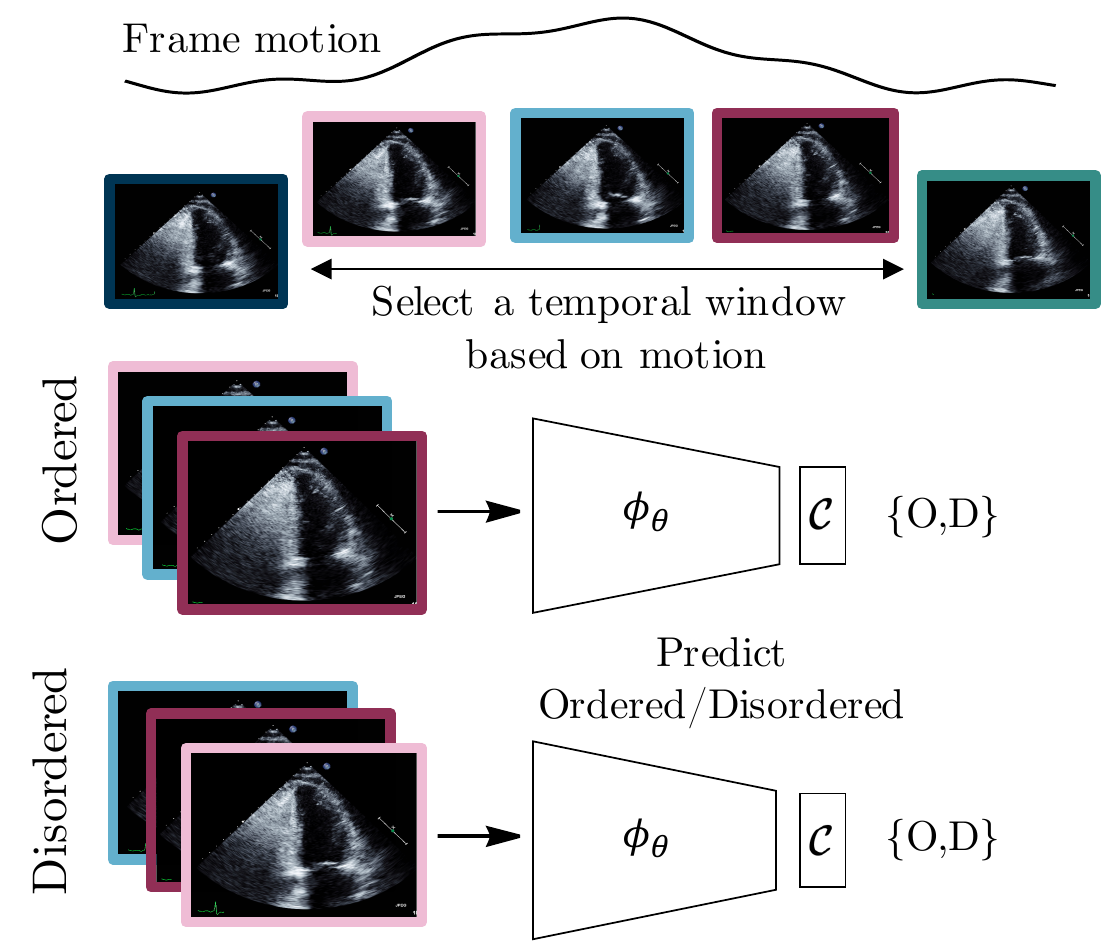}
\end{center}
\caption{Temporal Intra-view Self-supervision: three-Tuples of frames are sampled from high motion windows in a cardiac cycle. A small binary classifier $\mathcal{C}$ is trained end-to-end with the encoder $\phi_\theta$ to identify sequences as being ordered or disordered. It should be noted that this task is applied to all cine views separately; here, we show the AP2 view as an example.}
\label{fig:temporal_intra}
\end{figure}

In sequential data types, such as echo cines, the ordering of elements encodes an easily understandable notion of the temporal structure. To exploit this property, it is natural to consider the self-supervised task of verifying if cines are arranged in correct chronological order. To frame this problem as a direct supervision over the quality of the embeddings, we implement a binary classifier $\mathcal{C}: \mathbb{R}^{3,d}\mapsto \mathbb{R}$ that attempts to classify sequences of three unique $d$-dimensional feature vectors (three-tuples) output by $\phi_\theta$ as either appearing in sorted or unsorted order in the original cine. Inspired by~\cite{Misra2016ShuffleAL} to make sure that the classification of the tuples is not ambiguous, we only sample tuples from temporal windows with high motion. As shown in ~\autoref{fig:temporal_intra}, a coarse optical-flow is used as a measure of motion between frames, and the training data for $\mathcal{C}$ is chosen by considering the motion as a sampling weight---frames with higher motion have a higher probability to be selected. Within a selected temporal window, all possible sorted three-tuples are selected as positive training examples and unsorted three-tuples as negatives. A sorted tuple may follow increasing or decreasing temporal order while an unsorted tuple follows neither. This simple sequential verification task encourages the embeddings to encode information about the ordering of events in a cardiac cycle.

To function as a direct supervision over the embeddings, the joint network ($\phi_\theta \circ \mathcal{C}$) is trained end-to-end. The parameters of $\phi_\theta \circ \mathcal{C}$ are learned by minimizing the binary cross-entropy loss of the predictions on each tuple. For a training sample $n$ this loss is defined as:
\begin{equation}
    \mathcal{L}_{\text{temporal-intra}}=\left(1-o_n\right) \log \left(1-\hat{o}_n\right)+o_n \log \left(\hat{o}_n\right),
    \label{tmpint}
\end{equation}
where $o_n\in\{0,1\}$ is a binary label indicating if sample $n$ is sorted and $\hat{o}_n\in [0,1]$ is the predicted probability that sample $n$ is sorted. The average value of this loss is computed over a given batch of data during training.

\subsubsection{Spatial Intra-view Self-supervision} 
We now formulate a self-supervised learning objective to help Echo-SyncNet understand the spatial similarities between nearby frames. We adopt a method originally proposed in \cite{npairloss_NIPS2016} in the context of distance metric learning, which is the process of learning embeddings that preserve the distance between similar data points. 

\begin{figure}[t]
\begin{center}
\includegraphics[width=.9\columnwidth]{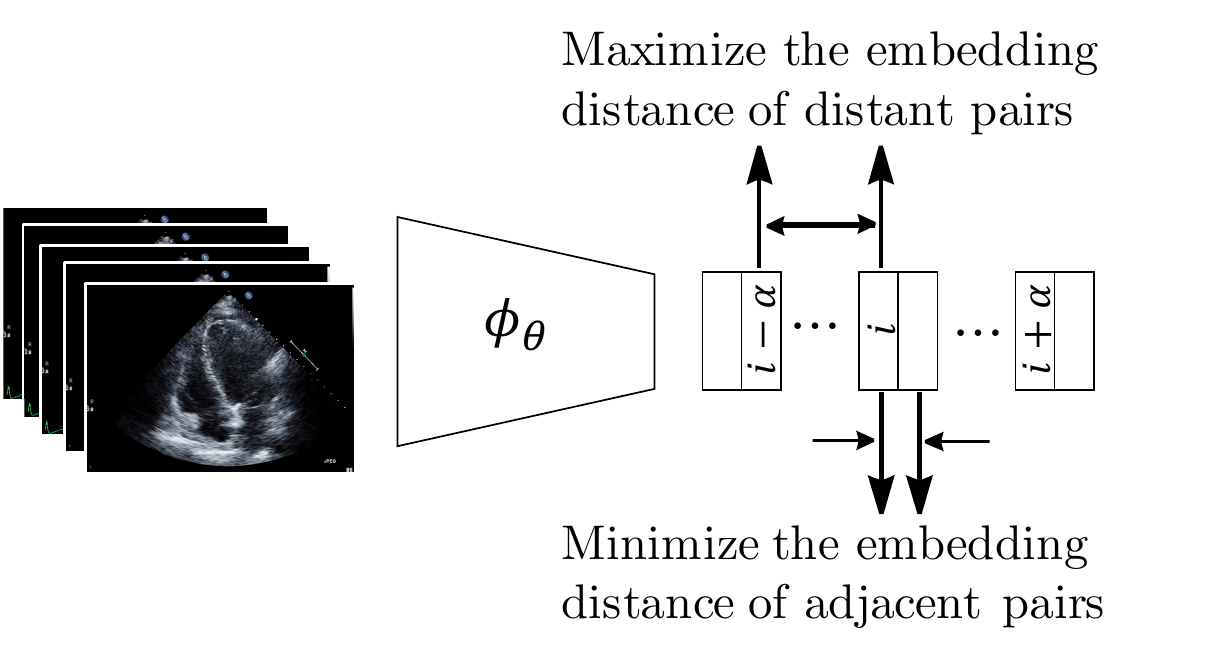}
\end{center}
\caption{Spatial Intra-view Self-supervision: Adjacent frames are selected for their embedding distance to be minimized, while distant frames (separated by at least $\alpha$) are selected for their embedding distance to be maximized. It should be noted that this loss function is applied to both cine views separately; here, we show the AP4 view as an example.}
\label{fig:spatial_intra}
\end{figure}

In our method, we partition the embeddings of an $n$ frame cine $\mathcal{Q}=\{q_1,q_2,\ldots,q_{n-1},q_{n}\}$ into positive and negative inner product pairs. For a given embedding $q_i$, there is a single positive pair $q_i^\top q_{i+1}$. The negatives are the inner products of all embeddings pairs separated by a minimum window radius $\alpha$---the window radius is a hyperparameter of our experiments further described in \autoref{implementation}. The objective is then formulated as minimizing the mean cross-entropy loss on all sequences of inner products of positive and negative pairs with the positive for each embedding as the target.
This loss function, further illustrated in \autoref{fig:spatial_intra}, is adapted from the $n$-tuplet loss originally formulated in \cite{npairloss_NIPS2016} and defined for this work in \autoref{ntup}. The intuitive objective of this loss is to maximize the inner product between adjacent elements of $\mathcal{Q}$ while minimizing it for distant elements:

\begin{equation}
\begin{split}
    \mathcal{L}_{\begin{subarray}{l}\text{inter}\\\text{spatial}\end{subarray}} &= \frac{1}{n-1} \sum_{i=1}^{n-1}-\log\left( \frac{e^{\nu_i}}{e^{\nu_i}+\sum_k e^{\mu_i^k}} \right)\\
    &=\frac{1}{n-1} \sum_{i=1}^{n-1} -\nu_i +\log\Big(e^{\nu_i}+ \sum_{k}e^{\mu_i^k} \Big) 
\end{split},
\label{ntup}
\end{equation}
where $\nu_i$ and $\mu_i^k$ are the positive and negative inner product pairs shown in \autoref{alpha} and $n$ is the number of frames in the cine for which this loss is computed on:

\begin{equation}
\begin{split}
        &\nu_i=q_i^\top q_{i+1}\quad \mu_i^k=q_i^\top q_{i+k}\\
        &\text{s.t abs}(k)\geq \alpha \text{ and }1\leq i+k \leq n 
\end{split}.
\label{alpha}
\end{equation}

This learning objective enforces that nearby and hence spatially similar frames map to similar embeddings and frames from different points of the cardiac cycle map to distant embeddings. For this loss to function properly, it is required that the training data consists of no more than one cardiac cycle. The reasoning for this restriction is to avoid pushing apart embeddings that correspond to the same cardiac event but are captured from different cycles.  

\subsection{Inter-view Self-supervision}
In this section, we describe the main framework of Echo-SyncNet for synchronizing multiple cardiac view cines. Given two echo cines from AP4 and AP2 views with $N$ and $M$ frames, respectively, their embeddings are noted as $\mathcal{P} = \{p_1, p_2, ...,p_N \}$ and $\mathcal{Q} = \{q_1, q_2, ...,q_M \}$. To formulate a self-supervised learning objective over a pair of embeddings ($\mathcal{P}, \mathcal{Q}$), we exploit their inherent consistency. Namely, the consistency property is that for each element $p_i \in \mathcal{P}$, its most similar element $q_j \in \mathcal{Q}$ should itself be most similar to $p_i$ out of all possible choices of $\mathcal{P}$. More plainly put, each distinct point in a cardiac cycle from one cine should correspond directly to the same point of the cardiac cycle in another cine. If the embedding sequences $\mathcal{P}$ and $\mathcal{Q}$ both come from the same patient and are each sampled over a single cardiac cycle, then upholding the consistency property is a valid objective for promoting the sharing of common features between embeddings.

Training a network to uphold the consistency property has two major pitfalls. First, due to the discrete nature of the definition, a pair of sequences can either satisfy the consistency property or not with no concept of how close a pair is to being consistent. Second, the property demands a bijective mapping between the elements of each sequence hence if the number of elements of the two sequences is not equal, they cannot be consistent. To address both of these problems, we adopt a differential loss function utilising a soft nearest neighbour approach as in \cite{TCC2019} that is minimized by making two sequences more consistent with one another.

For each embedding, $p_i\in\mathcal{P}$, its soft nearest neighbour $\tilde{q}_i\in\text{Span}(\mathcal{Q})$ is defined by \autoref{snn}:

\begin{equation}
\tilde{q}_i = \sum_j^M \frac{\exp(-||p_i - q_j||^2)}{\sum_k^M \exp(-||p_i - q_k||^2)} \ q_j .
\label{snn}
\end{equation}

Using $p_i$'s soft nearest neighbour $\tilde{q}_i$, we define $\beta_i^k$ as the normalized similarity between each $p_k \in P$ and $\tilde{q}_i$:
\begin{equation}
    \beta_i^k = \frac{\exp(-||\tilde{q}_i - p_k||^2)}{\sum_j^N \exp(-||\tilde{q}_i - p_j||^2) } .
\end{equation}

\begin{figure}[t]
\begin{center}
\includegraphics[width=\columnwidth]{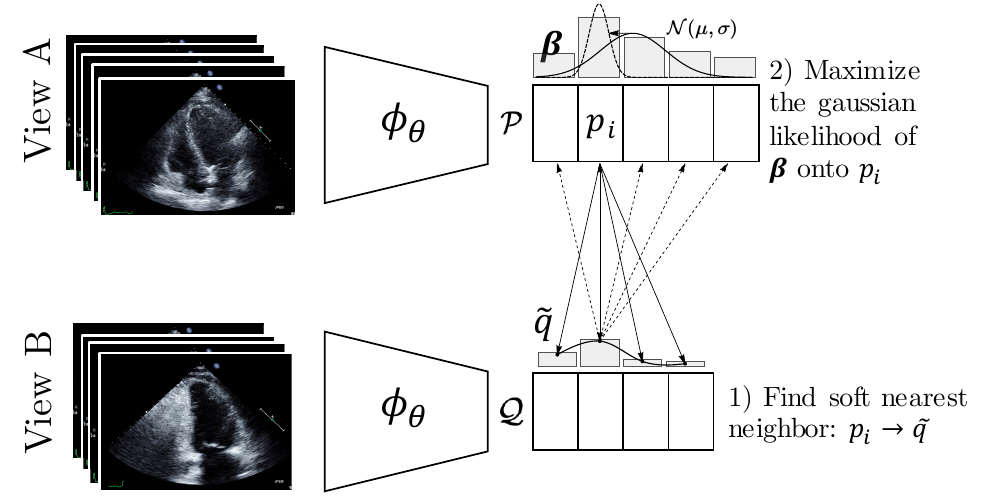}
\end{center}
\caption{Inter-view Self-supervision: Two echo cines taken in different views are encoded as two embedding sequences $\mathcal{P}$ and $\mathcal{Q}$. For each embedding in $p_i\in \mathcal{P}$, the soft nearest neighbor $\tilde{q}_i$ from $\mathcal{Q}$ is computed and compared with each element of $\mathcal{P}$ to make the discrete distribution $\boldsymbol\beta_i$. The encoder $\phi_\theta$ is trained to minimize the empirical negative log likelihood of the Gaussian prior $\mathcal{N}(\mu, \sigma)$ on $\boldsymbol\beta_i$ to the target position $i$.}
\label{fig:inter}
\end{figure}

To satisfy the the consistency property, we require that $\boldsymbol\beta_i=\{\beta_i^1,\ldots,\beta_i^N\}$ is maximized near the $i^{th}$ index. The final expression for the loss is formulated as in \cite{TCC2019} by imposing a Gaussian prior $\mathcal{N} (\mu_i, \sigma_i^2)$ defined using the empirical mean and variance of $\boldsymbol\beta_i$. The objective is then to minimize the negative log likelihood as shown in \autoref{nll} and \autoref{fig:inter}.
\begin{equation}
\begin{split}
     \mathcal{L}_{\mathrm{inter-view}} &=\frac{|i-\mu_i|^2}{\sigma_i^2} +  \frac{\lambda}{2} \log(\sigma_i^2), \\
     & \mathrm{where} \  \mu_i = \sum_{k=1}^N k \beta_i^k \ \ \mathrm{and} \ \  \sigma_i^2 = \sum_{k=1}^N \beta_i^k(k-\mu_i)^2 .
\end{split}
\label{nll}
\end{equation}

The regularization $\lambda$ is an additional hyperparameter to the loss that will be further described in \autoref{implementation}.

\section{Experiments and Results}

\subsection{Dataset}
\label{dataset_sec}

The echocardiography dataset used in this work was collected from the picture archiving and communication system (PACS) server of the Vancouver General Hospital with ethics approval from the Institutional Medical Research Ethics Board in coordination with the Information Privacy Office. Our first of three sets of experiments are carried out on a dataset consisting of 998 patient studies using a mutually exclusive train and test split of 854 and 144, respectively. As each patient may have multiple echos, this dataset contains a total of 1996 AP4 and AP2 pairs. No explicit temporal correspondences between cines are provided in this dataset.

The second set of experiments focuses on a direct comparison with a supervised learning approach for fine-grained cardiac cycle phase detection in AP4 cines. This set of experiments was conducted using data from 3070 patients, further details of which are explained in \autoref{ablation} under our ablation study.

The third set of experiments focuses on the robustness and  generalizability of the learned embeddings. This set of experiments was conducted using PLAX, AP4 and AP2 views taken from 1508, 2355 and 2220 patients respectively. All studies were obtained under the same ethics approval. Further details of which are explained in \autoref{extension} and \autoref{one_shot}. The datasets used in both the second and third sets of experiments contain labeled keyframes for both end-diastole (ED) and end-systole (ES) phases of the cardiac cycle. These labels were assigned by trained sonographers as a part of routine clinical care using both the echo cine and the ECG signal. 

For preprocessing, echo cines were cleaned of user interface elements by applying a binary mask around the beam region. The pixels outside of the mask location are set to zero to make sure that the model is not biased to any annotations around the beam. Moreover, echo cines were downsized to a spatial dimension of $224 \times 224$. For the n-tuplet and synchronization loss to function correctly, it is required that each cine in our training set contains approximately one cardiac cycle. To enforce the one cycle constraint, we either used the provided average heart rate and frame rate data to approximate a single period, or in cases where this information was not available, we applied the class agnostic video repetition counting method (RepNet) proposed in~\cite{Dwibedi_2020_CVPR}. In training studies with more than one cycle, all available cycles are considered as individual training samples.

\subsection{Implementation Details}
\label{implementation}

The encoder network $\phi_\theta$ is composed of two main modules. The first is a base 2D convolutional network that acts on each frame individually. A ResNet-50 \cite{ResNet2016} backbone truncated after the Conv4c layer is used as the architecture for the base network. Down-sampling in ResNet-50 is done using convolutions with stride 2. The dimension of the last feature map of the base network is $14\times14\times1024$. For each frame's last feature map, $k$ adjacent maps in a context window are stacked along the temporal dimension and passed to the second part of the encoder. We empirically chose $5$ as the size of the context window. The second part of the encoder is a 3D convolutional network designed to aggregate temporal information~\cite{dezaki2019frame}. The structure of this network is two 3D convolution layers followed by 3D max-pooling and a fully connected layer to output a 128-dimensional embedding vector. The encoder produces one embedding vector for each frame in the input cine.

 For the temporal intra-view self-supervision loss, we sampled eight triplets per echo cine from which six of the triplets were shuffled. Experiments showed that it was important to have a larger percentage of disordered triplets in our training set. The classifier ($\mathcal{C}$), described in \autoref{temp-intra}, takes three 128-dimensional vectors as input and consists of a concatenation layer followed by two fully connected layers of output dimension $128$ and $64$ with ReLU activation and a final linear layer of output dimension two produces logits for binary classification.

For the spatial intra-view self-supervision loss, the window radius $\alpha$ for selecting distant embedding pairs was set to $5$. The regularization weight $\lambda$ in the inter-view loss defined in \autoref{nll} was set to $0.001$.

When evaluating the inter-view loss on a pair of cines we first select 20 uniformly random frames from each cine to be used at inputs. When then compute the per frame embeddings of each set of 20 frames using $\phi_\theta$ at which point we can come compute the inter-view loss with \autoref{nll}. Using a fixed number of input frames allows for simpler and more efficient forward passes and computations of the loss on large batches of data in parallel. Standardizing frame number by selected frames randomly showed superior to using temporal interpolation and inherently makes our method more robust to variations in framerate.

The three self-supervised loss functions were linearly combined to form a single objective. To determine the optimal weights of each loss term, we performed a grid search over the range $[0,1]$ with grid size of $0.25$. The decided weights were $0.25$, $0.25$, and $0.5$ for the temporal intra-view, spatial intra-view and inter-view self-supervisions, respectively. Our models were trained using an Adam optimizer with a batch size of $4$. The initial learning rate was set to $10^{-4}$ with a scheduled decay rate of $0.1$ after every $5000$ iterations.

\subsection{Qualitative Results}
To synchronize two echo cines, we first find their self-supervised representations using the encoder network $\phi_\theta$. Next, the correspondence of each frame across cines is found using a dynamic time warping method \cite{dtw}, which is an algorithm for time series alignment. 

\begin{figure*}[t]
\begin{center}
\includegraphics[width=.8\textwidth]{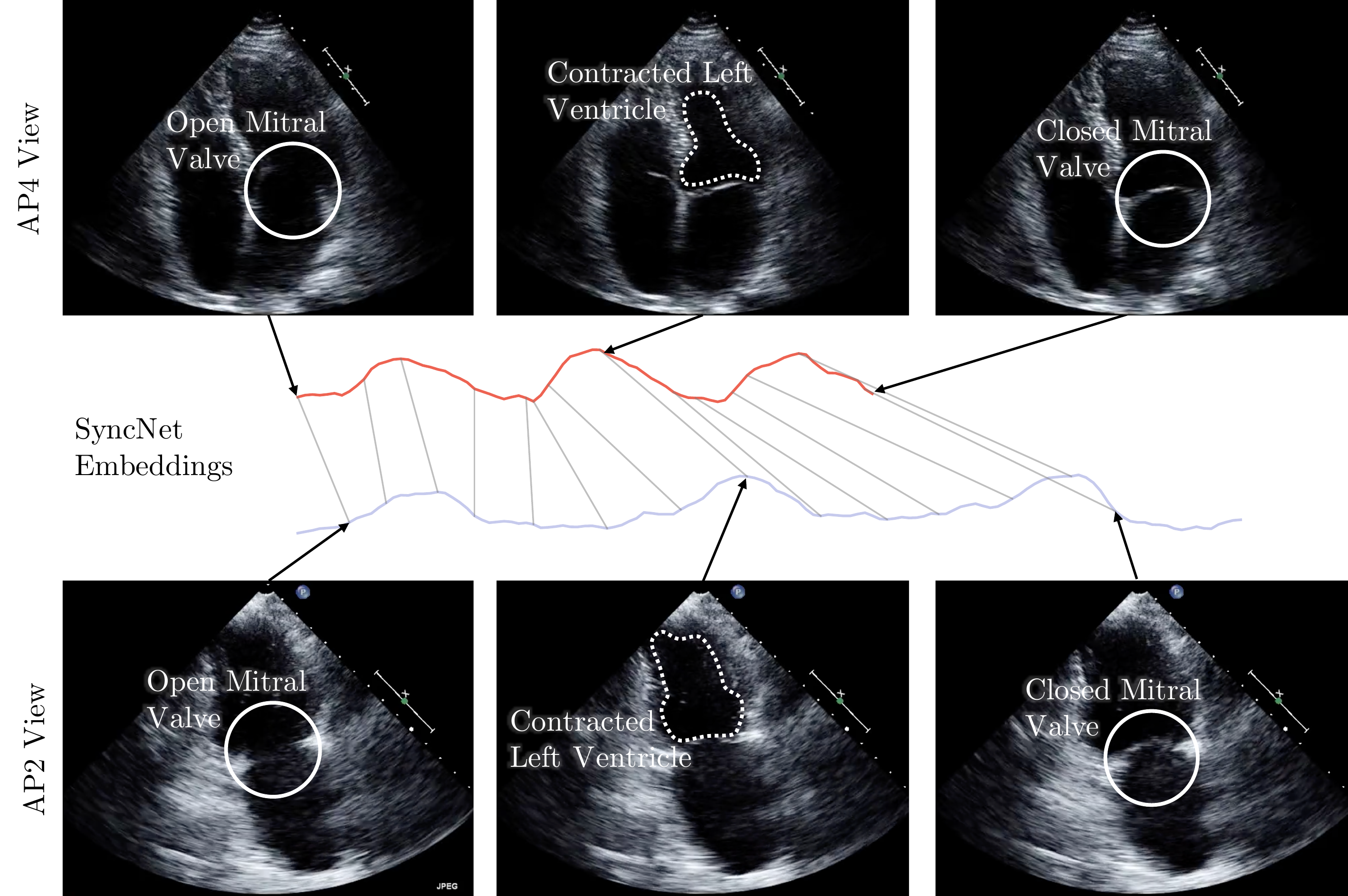}
\end{center}
\caption{A sample synchronization produced by Echo-SyncNet. We examine the synchronization of an AP4 and AP2 echo cine by comparing three distinct cardiac events: the earliest opening of the mitral valve, the maximum contraction of the left ventricle and the earliest closing of the mitral valve. We show the correspondence of these three cardiac events on the respective embeddings generated by Echo-SyncNet --- AP4 (red), AP2 (blue). Additionally, a full mapping between the AP4 and AP2 embeddings is shown with thin gray lines. Please see the supplementary files for synchronized video results.}
\label{fig:visual-result}
\end{figure*}

\begin{figure}
    \centering
    \includegraphics[width=\linewidth]{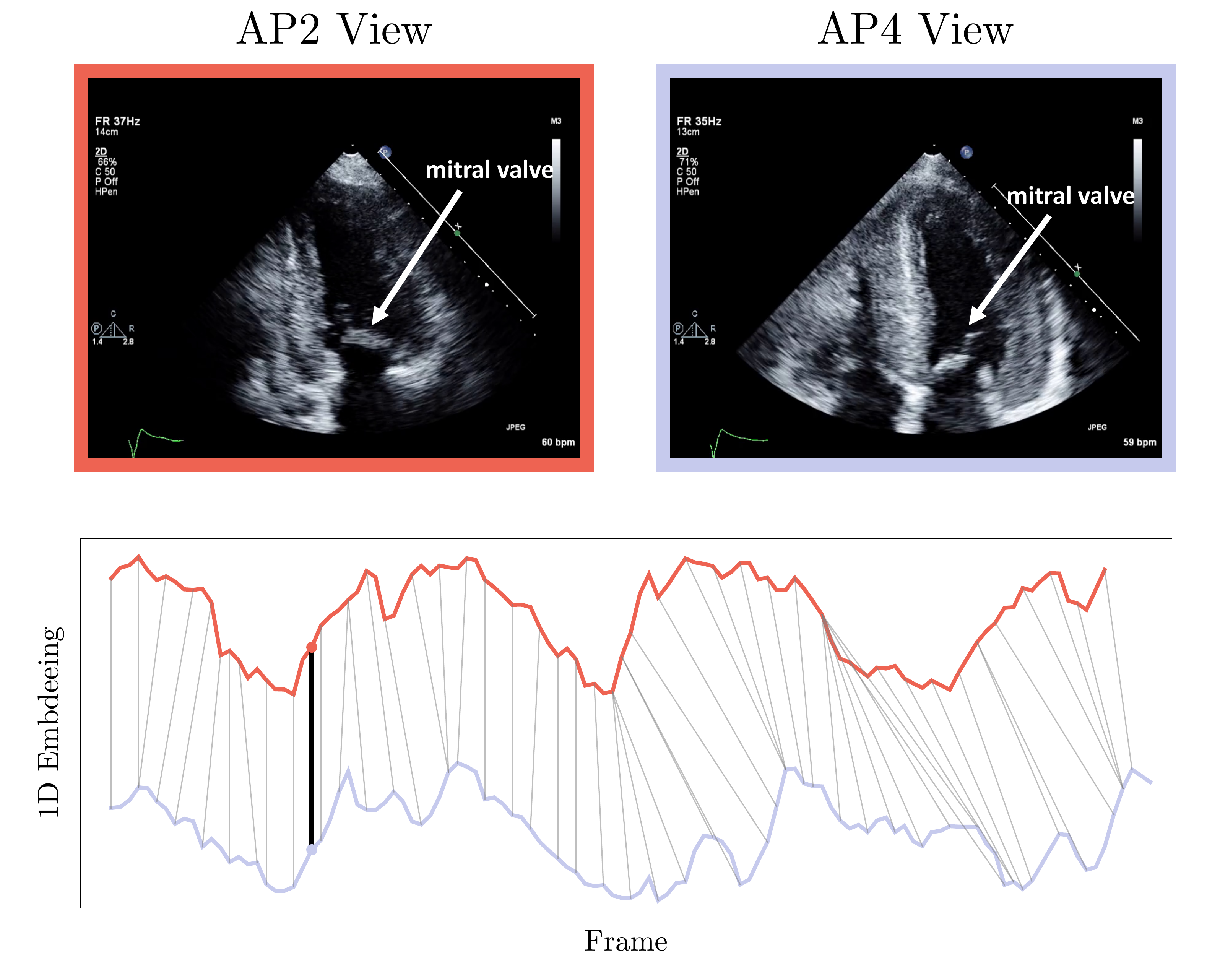}
    \caption{Visualization of embeddings: For each view, we show trajectories of the cines in the embedding space. Using a principal component analysis reduction approach, we reduce the dimensionality of the embedding from 128 to 1 for visualization. The visualization indicates how the embeddings change as the cardiac cycle progresses and how two cines with a different number of frames can be synced together. The images show a sample correspondence point (shown with black line in the lower sub-figure) for an AP4 and AP2 cardiac echo as well as their respective embeddings. Notice the corresponding position of the mitral valve in both images, and for a more obvious correspondence of cardiac walls and chambers during a cardiac cycle please see the supplementary videos.}
    \label{fig:detailed_embs}
\end{figure}

In \autoref{fig:visual-result}, the synchronization of an AP4 and AP2 echo cine is shown. To examine the details of the synchronization keyframes showing the max contraction of the left ventricle and the opening/closing of the mitral valve are shown. Additionally, we show the dynamic time warping correspondence between the embeddings of the two cines. Further information on the visualization of the cine embeddings in \autoref{fig:visual-result} and the supplementary material is explained in \autoref{fig:detailed_embs}.  


\subsection{Empirical Ablation Analysis}
\label{ablation}

We evaluated the learned representations with two frequently used metrics in prior literature to assess temporal synchronization of two videos~\cite{dezaki2018cardiac,dezaki2017deep,kong2016recognizing, TCC2019}, Kendall's Tau~\cite{kendalls_tau} and the coefficient of determination ($R^2$ score).
To automatically estimate the synchronization quality between cines, we measured Kendall's Tau for models trained with different combinations of the three self-supervisions. Kendall's Tau ($K's \ \tau $) is a statistical metric that measures the ordinal association between two sequences by counting the number of pairwise concordances and discordances between elements. For every pair of frame embeddings in the first cine, $(p_i, p_j)$ where $i<j$, the nearest neighbour correspondence in the second cine $(q_{\alpha},q_{\beta})$ is computed. A matching is concordant if and only if $\alpha < \beta$ and is otherwise discordant. For a given pair of cines, Kendall's Tau is calculated as: 
\begin{equation}
 K's \ \tau  =\frac{2}{n(n-1)}(\text{\#concordances}-\text{\#discordances}) .
\end{equation}
The highest attainable score is $1$ and implies that every matching is concordant, while the lowest score is $-1$ and is attained only for a completely reversed synchronization. We average Kendall's Tau over all pairs of cines in the validation set and the results are reported in Table~\ref{tab1}. Using solely inter-view self-supervision outperforms the joint spatial and temporal intra-view task as it considers two cines together as opposed to just one. However, the best performance is achieved by combining inter and intra-view supervision, which shows the impact of their complementary effects to reduce the model from over-fitting to a particular loss. The inter/intra-view combination is our final proposed method and will be used for the remaining experiments.

\begin{table}[!t]
\caption{AP2 and AP4 Synchronization Performance Evaluation based on Kendall's Tau metric}\label{tab1}
\centering
\begin{tabular}{|c|c|}
\hline
Method & $K's \ \tau$  \\
\hline\hline
Temporal Intra-view (TIV) & 0.772 \\
\hline
Spatial Intra-view (SIV) & 0.694\\
\hline
Inter-view (IV)~\cite{TCC2019}& 0.801\\
\hline
IV  and TIV  & 0.784\\
\hline
 IV and SIV & 0.713\\
\hline
Echo-SyncNet (Proposed) & \textbf{0.921}\\
\hline

\end{tabular}
\end{table}

\smallbreak
Embeddings can carry semantic or structural meanings. To test the utility and robustness of our embeddings, we decided to take advantage of them in a practical downstream task, i.e. fine-grained cardiac phase detection. A typical cardiac cycle consists of two major phases, one during which all chambers expand and refill with blood, termed diastole, followed by a period of contraction and pumping of blood, termed systole. In ~\cite{dezaki2017deep}, authors assigned each frame a fine-grained phase value, which mimics the left ventricle volume using the following equation:
\begin{equation}
\label{eq:label}
y_t = 
\begin{cases}
     \left(\frac{|t-T_\text{ES}^\textit{i}|}{|T_\text{ES} ^\textit{i} -T_\text{ED}^\textit{i}|}\right)^{3}, & \text{if}\ T_\text{ED}^\textit{i}<t<T_\text{ES}^\textit{i} \\
      \left(\frac{|t-T_\text{ES}^\textit{i}|}{|T_\text{ES}^\textit{i}-T_\text{ED}^\textit{i + 1}|}\right)^{\frac{1}{3}}, & \text{if}\ T_\text{ES}^\textit{i}<t<T_\text{ED}^\textit{i+1} .
\end{cases}
\end{equation}

 \begin{figure}
    \centering
    \resizebox{\columnwidth}{!}{%
    \includegraphics[width=0.5\textwidth]{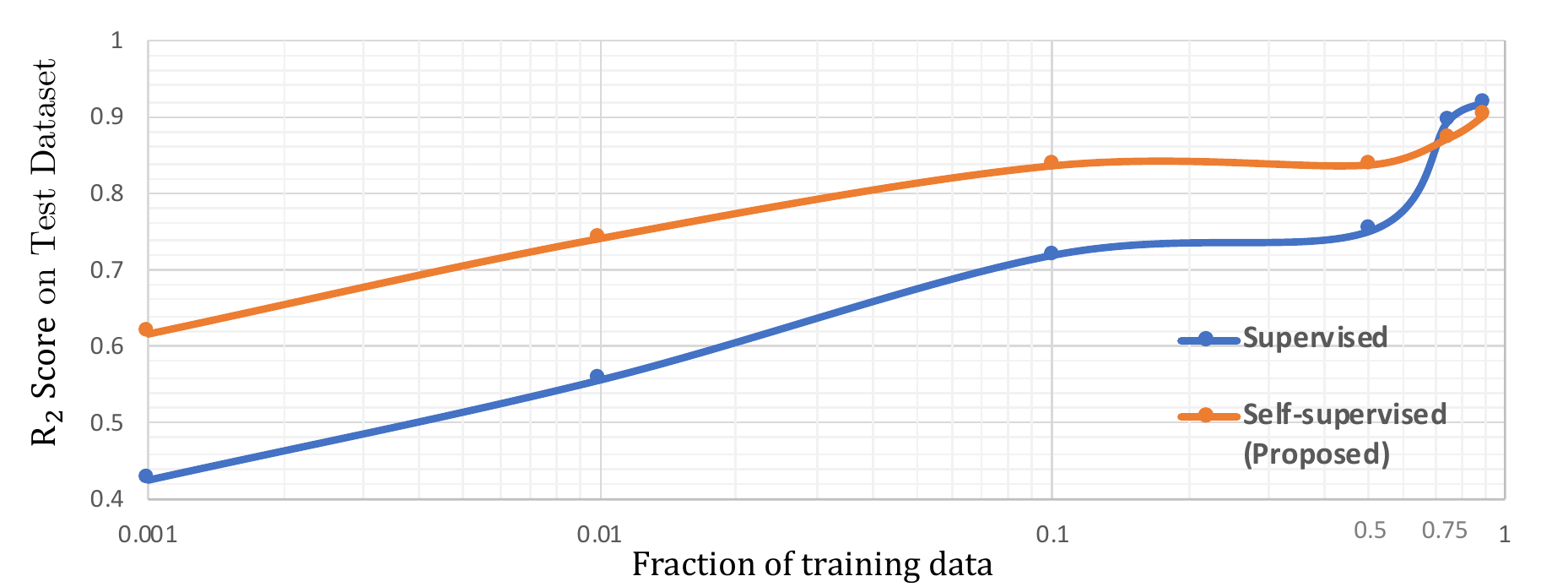}%
    }
    \caption{Comparison of self-supervised learning with supervised learning in fine-grained phase detection with different fractions of training data.}
    \label{fig:chart_result}
\end{figure}

Where $t$ is the frame index within cardiac cycle $i$, $T_\text{ED}^i$ is the index of end-diastolic frame in cardiac cycle $i$, and $T_\text{ES}^i$ is the index of end-systolic frame cardiac cycle $i$. Using this equation, each frame in a cardiac cycle is assigned to a phase value between 0 and 1. These values show an approximate measure of progress through each phase. To test the representation power of our embeddings, we used a linear regression to try and estimate the phase values directly from the embedding values. We carried out this experiment on a large set of AP4 cines from 3070 individual cases. We partitioned the data into mutually exclusive sets of training (60$\%$),  validation (20$\%$) and test (20$\%$) based on unique patients. No patient studies are shared across these sets. All the results shown were computed on unseen test dataset. Ground truth values for $T_\text{ED}$ and  $T_\text{ES}$ were recorded by sonographers. Using the ground truths, the label for each frame was calculated by (\ref{eq:label}). The performance of this experiment was measured based on the coefficient of determination ($R^2$ score) which is defined by: 
$$R^2 = 1-\frac{\sum(y_t-\hat{y}_t)^2}{\sum(y_t-\bar{y})^2}~,$$ 
where $\bar{y}$ is the mean of true labels and $\hat{y}_t$ is the prediction for $y_t$. The $R^2$ score of the regression is reported in ~\autoref{fig:chart_result}. We also show the advantage of applying self-supervised learning as apposed to supervised training in scenarios with different fractions of labeled training data. Following the experts in this field \cite{TCC2019}, we train the embedding space either by the proposed self-supervised method or a vanilla supervised method. Then, the aforementioned linear regressor is trained on top of embeddings using different fractions of labeled training data. We observe significant improvement using self-supervised learning in experiments with a small amount of labeled data, from which can be inferred that the embedding space is generalizable for downstream tasks in low-labeled data scenarios. This shows that there is much untapped rich information present in the raw echo cines that can be harvested using self-supervision. A fraction of training data increases, the performance gaps narrow between supervised learning and self-supervised learning, with a slightly better performance in supervised learning as the model is explicitly optimized for the fine-grained phase-detection task. However, self-supervised approach is an effective learning strategy when training target tasks on small annotated data-sets. Such strategy has a significant potential for medical imaging analysis, as most of the time, there is a shortage of labeled data. It is not only difficult to get a large dataset labeled because of the manual effort involved, but also the labels may suffer from significant intra/inter-observer variability, specifically amongst less experienced operators~\cite{bresser2015study,cole2015defining}.

\subsection{Extension of Echo-SyncNet to Other Views}
\label{extension}

Although we experimented on the AP4 and AP2 views up to this point, the proposed deep learning approach does not make any \emph{a priori} assumption on a specific echo view.
We conduct a feasibility study with the use of 1508 patient studies on Parasternal Long Axis (PLAX) view, 2220 patient studies on AP2 and 2355 patient studies on AP4 obtained under the same ethics approval. We sample random pairs of cines from these three views and present each pair to the encoder network, $\phi_\theta$, to learn the inter- and itra-view self-supervisions.  The encoded embeddings capture semantic information from this new combination of views and can be used to synchronize them smoothly. Please refer to a sample synched video in the supplementary files to watch its performance. The embeddings can be used to synchronize more than two views, and as a sample, we provide a video in the supplementary files that shows AP2, AP4, AP5 and PLAX cines can be synchronized smoothly. 
This experiment can have significant promise for echo imaging, as it determines the breadth of network applicability over various captured echo views.

\subsection{One-shot Learning for Key Frame Detection}
\label{one_shot}
We conducted an experiment to test the usefulness of our learned embeddings in the one-shot scenario. Using only one reference cine with multiple cycles and labeled ED/ES keyframes, we test the prediction power of identifying key frames in unseen cines. The reference and a candidate cine are synchronized using Echo-SyncNet. Once the synchronization is found, the ED and ES frames are identified in the candidate cine based on their correspondence to the reference. Using the sonographer labeled ground-truth key frames for the candidate cine, the mean absolute error ($\mu_{ED}$, $\mu_{ES}$) and the standard deviation of the error ($\sigma_{ED}$, $\sigma_{ES}$) are computed for AP4, AP2 and PLAX views. As cines have been recorded at different frame rates, we report statistics in both the number of frames and milliseconds.

In Table~\ref{tab:edessync}, the results of key frame detection in three different views can be observed. Please note that for these predictions, only one labeled sample has been used for each view. The errors are similar to reported inter-observer variability for manual detection of ED and ES frames by sonographers (Zolgharni \emph{et al.}~\cite{zolgharni2017automatic} demonstrated that the median disagreement among five sonographers for the identification of ED and ES phases is $3$ frames).

\begin{table}[]
    \centering
    \caption{One Shot Learning Results for ED/ES Detection}
    \begin{tabular}{|l|l|l|ll|ll|}
    \hline
\multirow{2}{*}{View} & \multirow{2}{*}{\begin{tabular}[c]{@{}l@{}}Validation \\ Size\end{tabular}} & \multirow{2}{*}{ Units} & \multicolumn{2}{c|}{\begin{tabular}[c]{@{}c@{}}Mean \\ Absolute \\ Error\end{tabular}} & \multicolumn{2}{c|}{\begin{tabular}[c]{@{}c@{}}Standard \\ Deviation\end{tabular}} \\
                      &                                                                             &                   & $\mu_{ED}$                                         & $\mu_{ES}$                                        & $\sigma_{ED}$                                      & $\sigma_{ES}$                                    \\ \hline\hline
\multirow{2}{*}{AP4}  & \multirow{2}{*}{486}  & frames & 4.1  & 3.6  & 3.6 & 2.9                                     \\
                      && ms  & 138 & 121  & 120  & 98                                     \\ \hline
\multirow{2}{*}{AP2}  & \multirow{2}{*}{408}& frames& 6.8 & 8.0 & 4.2& 3.5                                     \\
                      & & ms & 136& 159 & 84& 70                                     \\\hline
\multirow{2}{*}{PLAX} & \multirow{2}{*}{294}  & frames   & 3.1  & 3.9  & 2.4  & 2.6                                     \\
                      &  & ms & 122  & 153 & 95 & 101 \\ \hline                
\end{tabular}
    \label{tab:edessync}
\end{table}

\subsection{Evaluation of the Proposed Method on an Unseen Public Dataset}

We performed an additional validation using a publicly available echo dataset, EchoNet-Dynamic~\cite{ouyang2020video}. The dataset includes 10,030 videos of AP4 echocardiography from patients who received imaging as part of routine clinical treatment at Stanford University Hospital between 2016 and 2018. The location of one ED and one ES frame is labelled in each video. We extracted the embeddings of these studies using our trained model on the multiview AP4, AP2 and PLAX dataset. The embeddings are validated in the detection of ED and ES frame location following the same procedure as the one explained in ~\autoref{one_shot}. In Table~\ref{tab:edesatanford}, the results of keyframe detection in this public dataset can be found. The errors are similar to the ones we got from our data, which is close to the inter-observer variability of ED and ES frame detection.

\begin{table}[]
    \centering
    \caption{One Shot Learning Results for ED/ES Detection in Publicly Available Echo Dataset (EchoNet-Dynamic) }
    \begin{tabular}{|l|l|l|ll|ll|}
    \hline
\multirow{2}{*}{View} & \multirow{2}{*}{\begin{tabular}[c]{@{}l@{}}Validation \\ Size\end{tabular}} & \multirow{2}{*}{ Units} & \multicolumn{2}{c|}{\begin{tabular}[c]{@{}c@{}}Mean \\ Absolute \\ Error\end{tabular}} & \multicolumn{2}{c|}{\begin{tabular}[c]{@{}c@{}}Standard \\ Deviation\end{tabular}} \\
                      &                                                                             &                   & $\mu_{ED}$                                         & $\mu_{ES}$                                        & $\sigma_{ED}$                                      & $\sigma_{ES}$                                    \\ \hline\hline
\multirow{2}{*}{AP4}  & \multirow{2}{*}{10,030 }  & frames & 4.0  & 3.1  & 6.7 & 4.6                                     \\
                      && ms  & 79.1 & 61.1  & 128.5  & 90.1                              \\ \hline                   
\end{tabular}
    \label{tab:edesatanford}
\end{table}

\section{Conclusion and Future Work}
In this paper, we proposed a self-supervised representation learning method for multi-view synchronization of echo cines. We showed that such rich yet free semantic representations can be used not only for synchronization, but also for fine-grained cardiac phase detection. The proposed technique does not make any \emph{a priori} assumption on a specific echo view and is hence performance insensitive to various cardiac views.

This work suggests that robust and generalizable feature representations can potentially be improved with self-supervision and opens a new direction towards finding novel self-supervision paradigms for robustness and generalizability rather than developing self-supervision approaches as a mechanism of catching up to fully-supervised performance.

In future, we plan to extend this method and foster such semantic representations from high dimensional cines in analysing the rhythm of the cardiac cycle and detecting abnormalities correlated with cardiac rhythm.

\section*{Acknowledgment}
The authors would like to thank contributions of the Natural Sciences and Engineering Research Council of Canada (NSERC) and the Canadian Institutes of Health Research (CIHR) for funding this project. We would like to also thank Dale Hawley for helping with accessing data used in this research.

\appendix

\section*{Description of Supplementary Material}

\begin{figure} [htb]
    \centering
    \includegraphics[width=\linewidth]{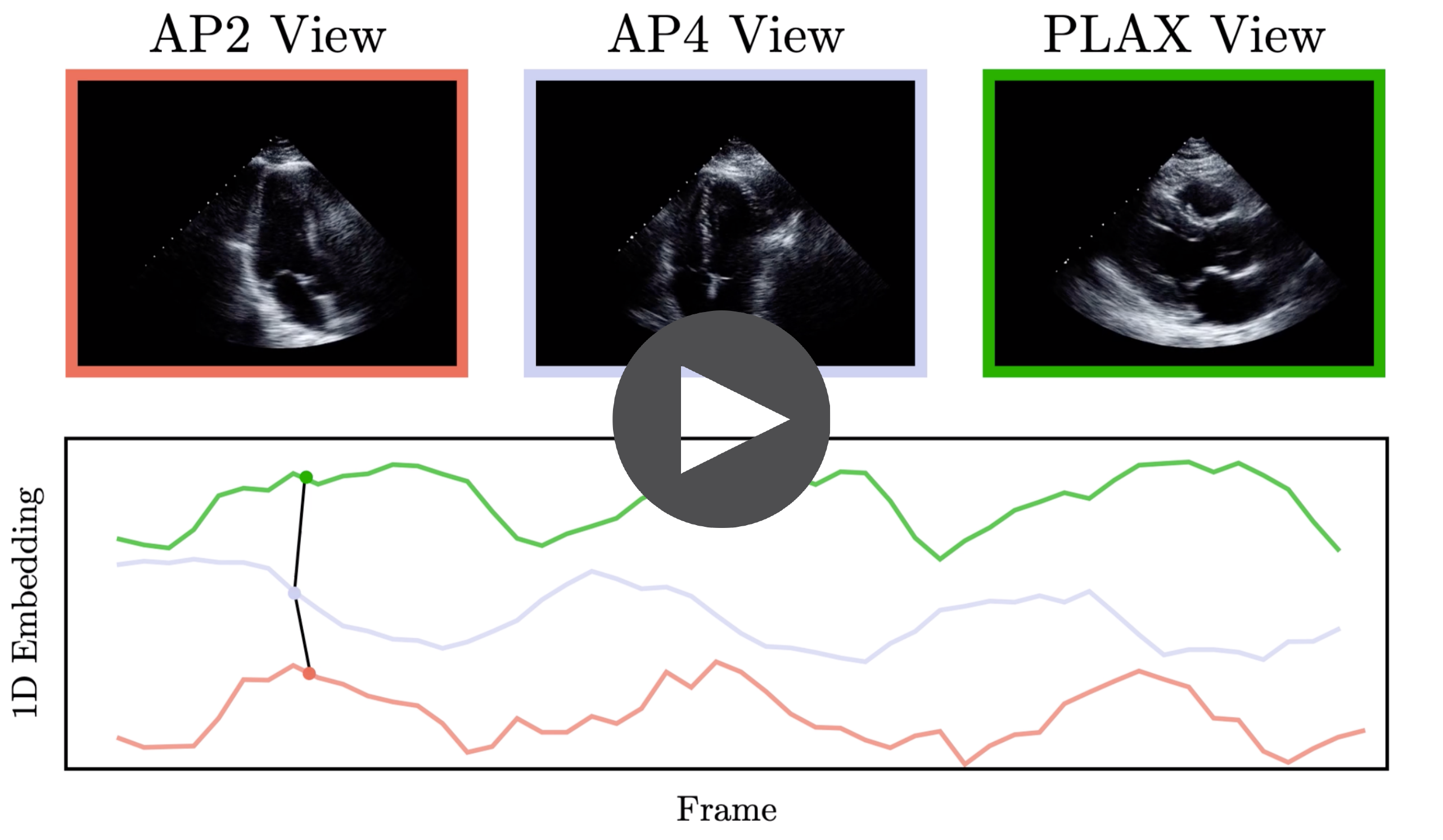}
    \caption{A screenshot of synched cines captured from AP4, AP2, and PLAX views along with their trajectories in the  embedding space. It should be noted that we reduced the embeddings' dimensions from 128 to 1 using principal component analysis for better visualization.}
    \label{fig:AP4_AP2_PLAX}
\end{figure}

\begin{figure}
    \centering
    \includegraphics[width=\linewidth]{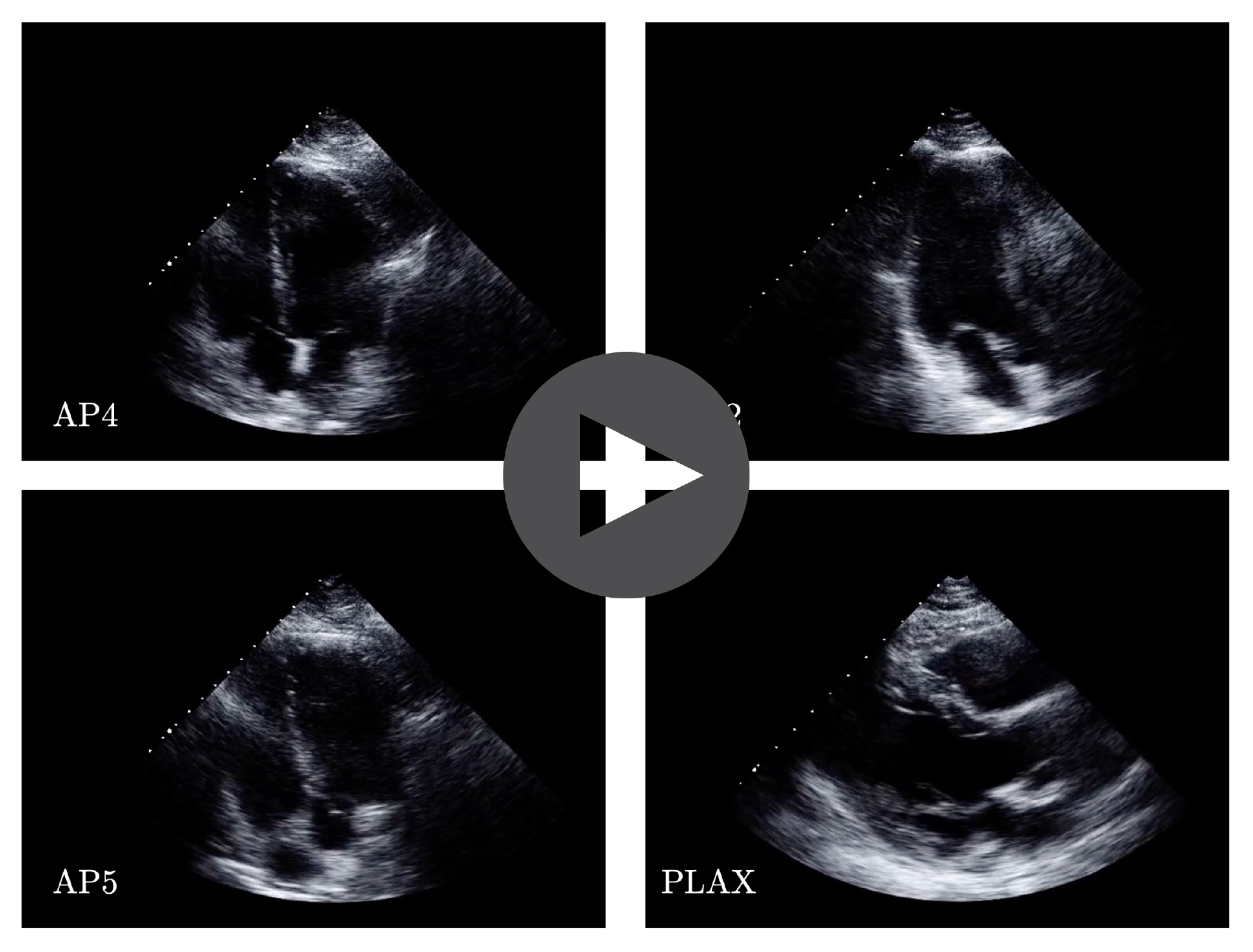}
    \caption{A screenshot of synced cines captured from AP4, AP2, AP5, and PLAX views.}
    \label{fig:AP4_AP2_AP5_PLAX}
\end{figure}

In ~\autoref{fig:AP4_AP2}, ~\autoref{fig:AP4_PLAX}, and ~\autoref{fig:AP4_AP2_PLAX}, we show a screenshot of synced cines captured from different views along with their trajectories in the  embedding space. Please refer to the supplementary files to see the videos. In synced videos, note the fine-grained synchronization that in some time points, frames are dropped or repeated to make the movements of valves and walls be aligned between the cines.
In ~\autoref{fig:AP4_AP2_AP5_PLAX}, a snapshot of a video from four synced cines can be observed. In this example, the generalizibility of the method even in the views not found in the training set can be observed.

\begin{figure} [htb]
    \centering
    \includegraphics[width=\linewidth]{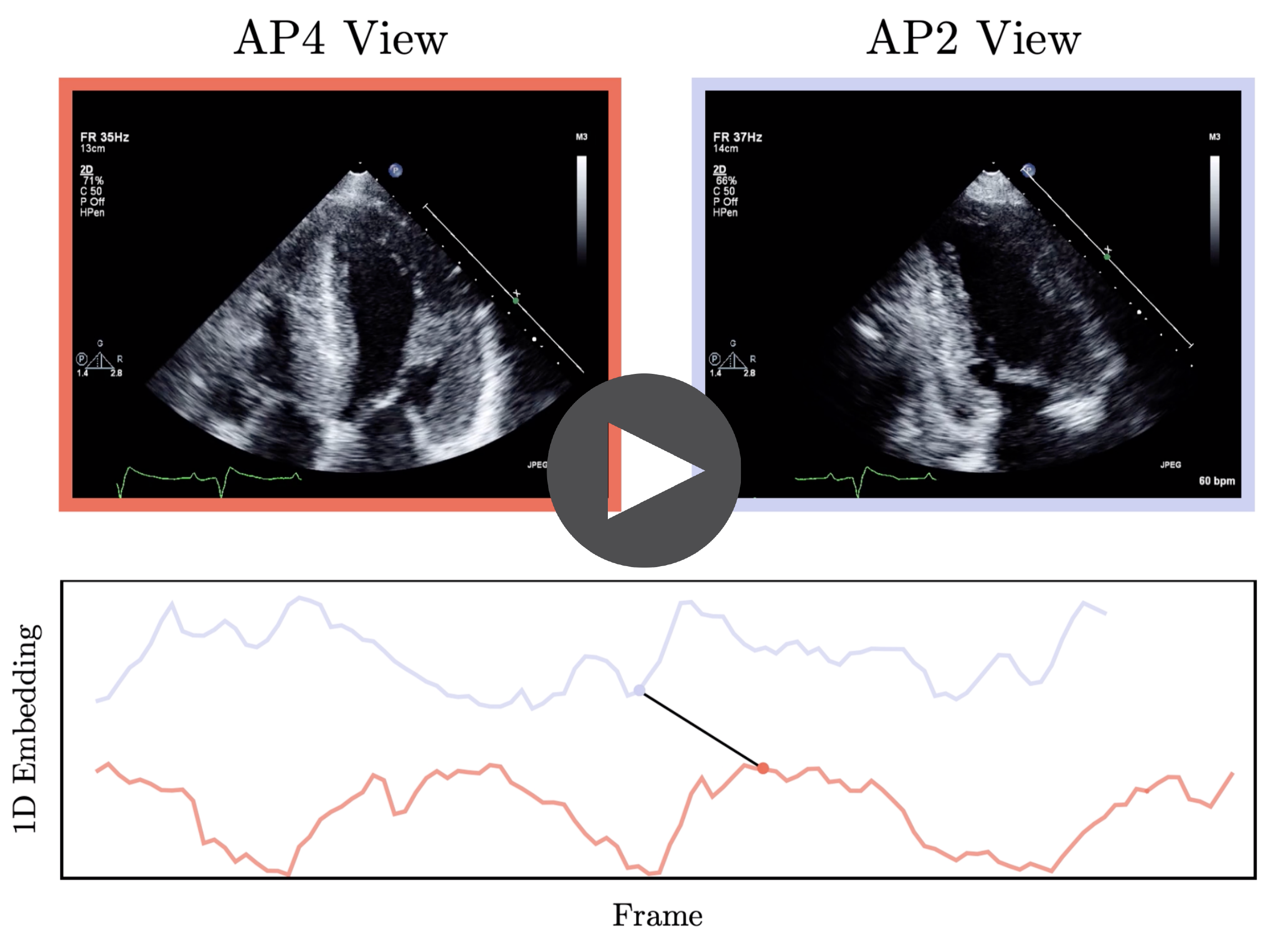}
    \caption{A screenshot of synched cines captured from AP4 and AP2 views along with their trajectories in the  embedding space . It should be noted that we reduced the embeddings' dimensions from 128 to 1 using principal component analysis for better visualization.}
    \label{fig:AP4_AP2}
\end{figure}

\begin{figure} [htb]
    \centering
    \vspace{-3mm}
    \includegraphics[width=\linewidth]{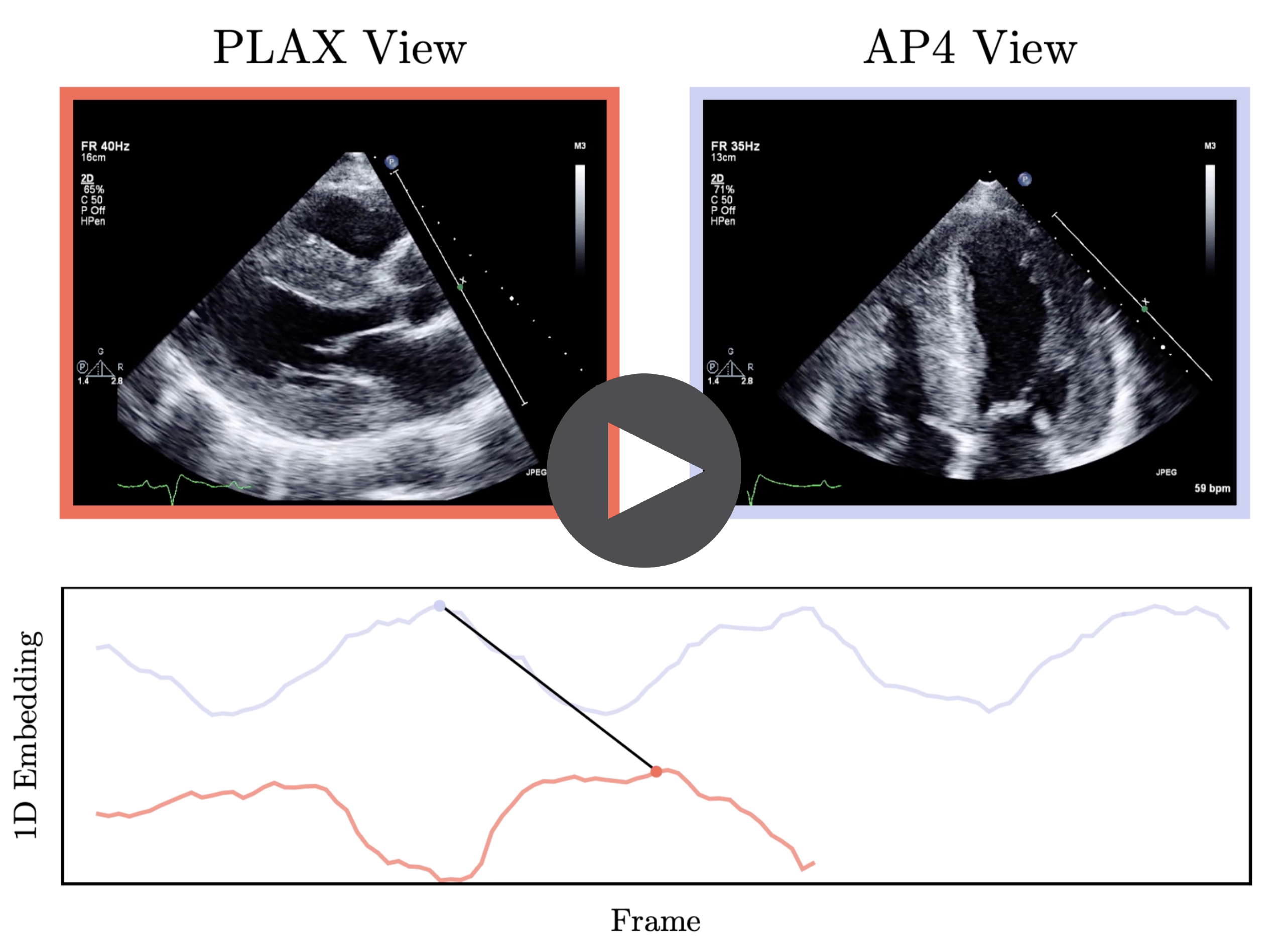}
    \caption{A screenshot of synched cines captured from AP4 and PLAX views, showing the extensibility and generalizability of the proposed model to other cardiac views. It should be noted that we reduced the embeddings' dimensions from 128 to 1 using principal component analysis for better visualization.}
    \label{fig:AP4_PLAX}
\end{figure}

\begin{figure*}[htb]
    \centering
    \includegraphics[width=.9\linewidth]{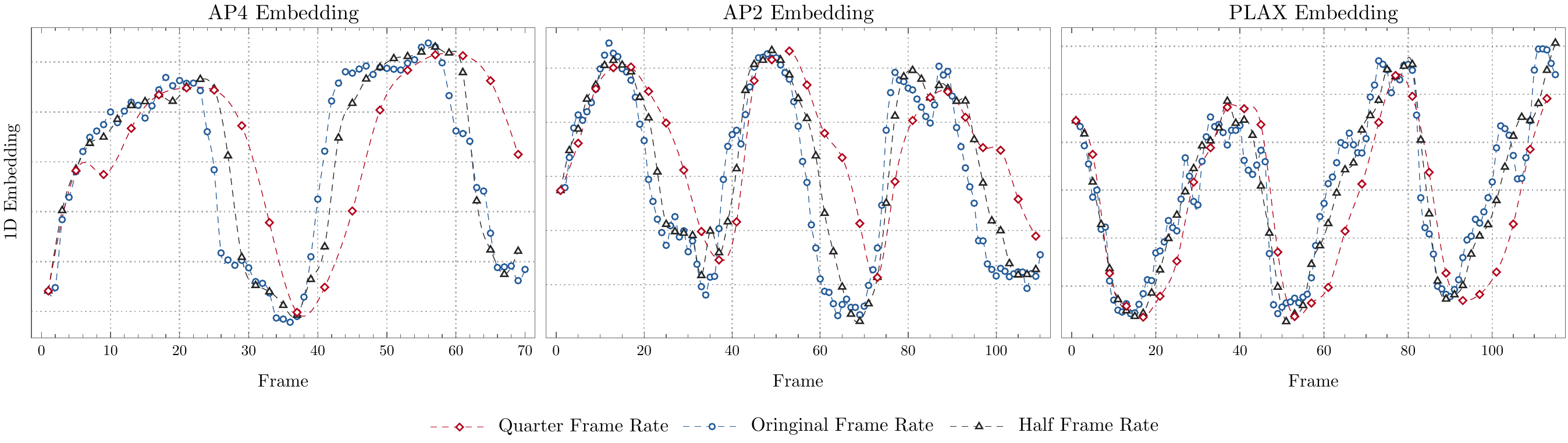}
        \caption{Qualitative sample of the sensitivity of embedding quality to frame rate.  Three sample cines of different cardiac views (AP4, AP2, PLAX) are selected. Each cine is sampled at three different frame rates (original, every second frame, every fourth frame). The 1D PCA reduction of each cine's embeddings is compared at different frame rates. The immediate visual conclusion is that even with a significantly lower input frame rate the resulting embeddings have a similar structure and will hence enable robust temporal alignment.}
    \label{fig:frsens}
\end{figure*}

\begin{figure} [htb]
    \centering
    \includegraphics[width=\linewidth]{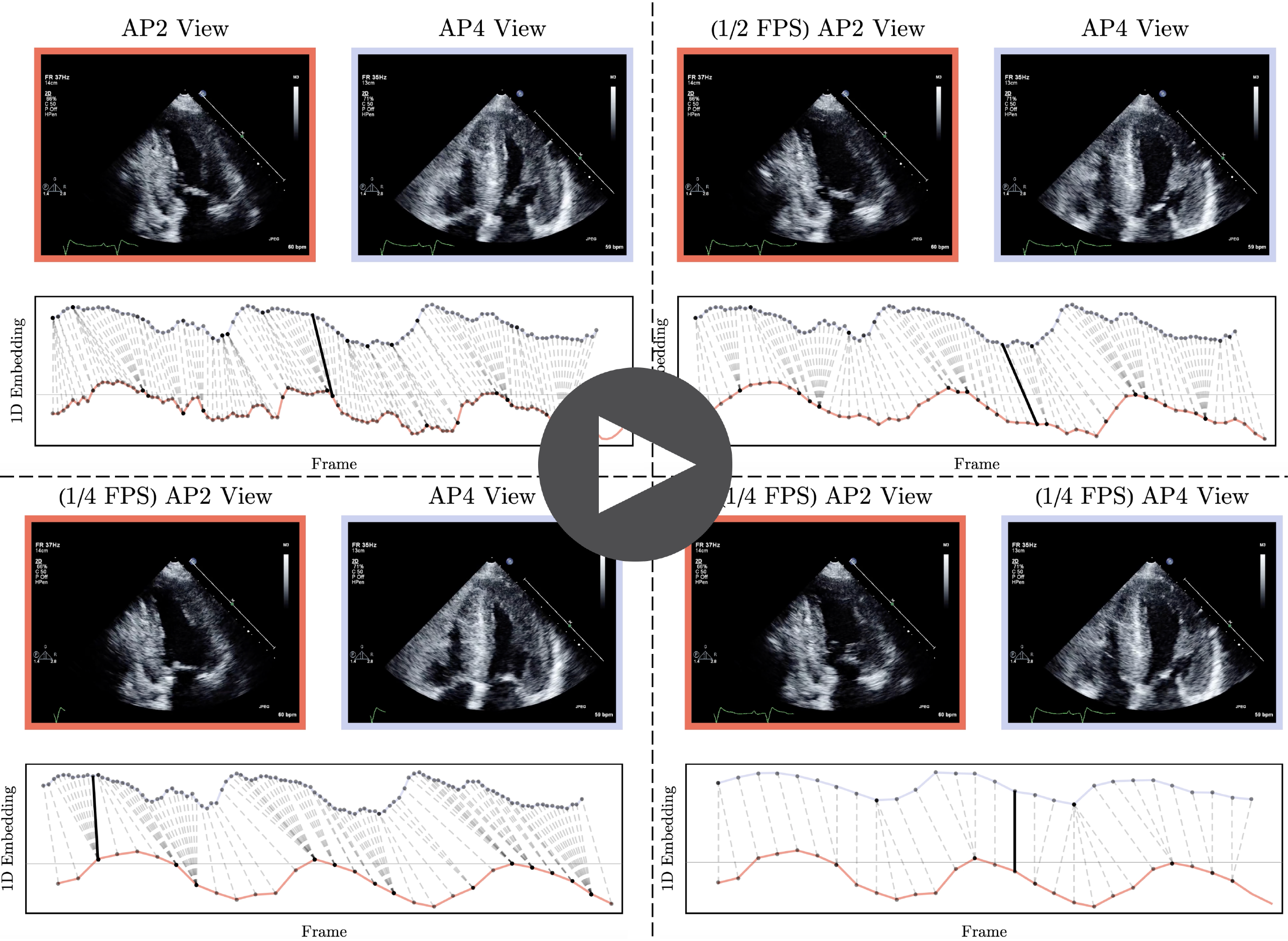}
    \caption{A screenshot of synched cines captured from AP4 and AP2 views along with their trajectories in the  embedding space. This experiment shows the robustness of the proposed synchronization method to different frame rates. We first produce downsampled versions of both cines at one-half and one-quarter of their original frame rate. Next, we create embeddings and perform synchronization using various combinations of original, half sampled and quarter sampled pairs. It should be noted that we reduced the embeddings' dimensions from 128 to 1 using principal component analysis for better visualization.}
    \label{fig:FR_sync}
\end{figure}

\section*{Hyperparameters}
We tabulate the list of values of the hyperparameters in Table~\ref{table_hyperparameters} .

\begin{table*}[!ht]
\caption{List of hyperparameters used}\label{table_hyperparameters}
\centering
    \begin{tabular}{|c|c||c|c|}
    \hline
    \textbf{Hyperparameter} & \textbf{Value} & \textbf{Hyperparameter} & \textbf{Value} \\
    \hline\hline
    Training Batch Size & 4& \thead{Temporal Intra-View Loss Weight\\($\lambda_1$ in \autoref{totalloss})} &0.25\\
    \hline
    Learning Rate & $10^{-4}$ & \thead{Spatial Intra-View Loss Weight\\($\lambda_2$ in \autoref{totalloss})} & 0.25\\
    \hline
    Learning Rate Decay & $0.1$ $/$ $5000$ batches & \thead{Inter-View Loss Weight\\($\lambda_3$ in \autoref{totalloss})} &0.5 \\
    \hline
    Optimizer & ADAM& \thead{Spatial Intra-View Window\\($\alpha$ in \autoref{alpha})} & 5 \\
    \hline
    \thead{Number of Input Frames\\During Training} & 20 & \thead{Fraction of Unsorted Samples for\\Temporal Intra-View Loss} &0.75\\
    \hline
    \thead{3D Convolution\\Context Window Size ($k$)} & 5 & \thead{Inter-View Loss Regularization\\($\lambda$ in \autoref{nll})} & $0.001$\\
    
    \hline

    \end{tabular}
\end{table*}

\section*{More Information on Datasets}
In an effort to reduce selection bias in the data used in this work, we include ultrasound studies from various machine vendors and do not perform filtering of data based on pathology or any other external variables. The unbiased nature of data selection should help the trained models to generalize for translation to real clinical settings. All data were captured as a part of routine clinical care at Vancouver General Hospital. More detailed information of the studies used in each of the experiments can be found  in~\autoref{tab:my_label}, ~\autoref{tab:my_label1} and \autoref{tab:my_label2}.



\begin{table*}[]
\caption{First Experiment (AP2-AP4 Synchronization)}
\hspace{1.5cm}
\begin{tabular}{|l|c|c|c|}
\hline
{} &   Heart Rate &   Frame Time (ms) & Pixel Width (mm)\\  
\hline\hline
mean  &    69.67 &    25.96 & 0.385\\\hline
std   &    16.49 &     7.42  & 0.071\\\hline
min   &    28 &     7.44  & 0.155\\\hline
25\%   &    59 &    20.09 & 0.332  \\\hline
50\%   &    67 &    23.29 & 0.384\\\hline
75\%   &    78 &    33.33 & 0.435 \\\hline
max   &   164 &    49.37 & 0.714 \\\hline
\end{tabular}
\quad 
\begin{tabular}{|l|c|}
\hline
     Year & \thead{Study Count}\\
\hline \hline
 2011 &   62 \\\hline
 2012 &   152 \\\hline
 2013 &   380 \\\hline
 2014 &  404 \\\hline
\end{tabular}
\quad
\begin{tabular}{|l|c|}
\hline
Machine & \thead{AP2-AP4 Study\\Count} \\
\hline\hline
SONOS & 9 \\\hline
Vivid i &  135 \\\hline
Vivid7 &  237 \\\hline
iE33 &  617 \\\hline\hline
\textbf{Total} & 998  \\\hline
\end{tabular}

\label{tab:my_label}
\end{table*}

\begin{table*}[]
    \centering
        \caption{Second Experiment (Fine-Grained Cardiac Cycle Phase Detection in AP4- captured with iE33)}

\begin{tabular}{|l|c|c|c|}
\hline
{} &   Heart Rate &   Frame Time (ms) & Pixel Width (mm)\\  
\hline\hline
mean &      68.40 &      25.56 &    0.365 \\\hline
std  &      12.98 &       7.21 &     0.065 \\\hline
min  &      35 &       9.06 &     0.192 \\\hline
25\%  &      60 &      19.84 &     0.332 \\\hline
50\%  &      67 &      22.05 &     0.358 \\\hline
75\%  &      76 &      33.33 &     0.395 \\\hline
max  &     141 &      44.60 &    0.767 \\\hline
\end{tabular}
\quad 
\begin{tabular}{|l|c|}
\hline
     Year & {Study Count}\\
\hline \hline
2010 &  246 \\\hline
2011 &  468 \\\hline
2012 &  304 \\\hline
2013 &  772 \\\hline
2014 &  1280 \\\hline

\end{tabular}
    \label{tab:my_label1}
\end{table*}

\begin{table*}[]
    \centering
        \caption{Third Experiment (Multi-View Synchronization)}

\begin{tabular}{|l|c|c|c|}
\hline
{} &  \thead{Heart\\Rate} &   \thead{Frame Time\\(ms)} & \thead{Pixel Width\\(mm)}\\  
\hline\hline
mean &      72.02 &      32.07 &  0.388 \\\hline
std  &      16.66 &       9.73 &  0.075 \\\hline
min  &      34 &      12.14 &  0.184 \\\hline
25\%  &      61 &      21.50 &  0.332 \\\hline
50\%  &      70 &      40.08 &  0.395 \\\hline
75\%  &      80 &      40.69 &  0.420 \\\hline
max  &     188 &      43.64 &  0.719 \\\hline
\end{tabular}
\quad 
\begin{tabular}{|l|c|}
\hline
Year & {Study Count}\\
\hline \hline
 2011 &   343 \\\hline
 2012 &   352 \\\hline
 2013 &   880 \\\hline
 2014 &  1304 \\\hline
 2015 &   248 \\\hline
 2017 &    1128 \\\hline
 2018 &  1828 \\\hline
\end{tabular}
\quad 
\begin{tabular}{|l|c|c|c|}
\hline
Machine & \thead{AP2 Study\\Count} & \thead{AP4 Study\\Count} & \thead{PLAX Study\\Count} \\
\hline\hline
SONOS  & 10 & 9 & 8 \\\hline
Vivid i  & 218 & 235 & 129 \\\hline
Vivid7  & 257 & 394 & 272 \\\hline
VividE95  & 0 & 0 & 540 \\\hline
iE33  & 1735 & 1717 & 561 \\\hline\hline
\textbf{Total} & 2220 & 2355 & 1508 \\\hline
\end{tabular}
\label{tab:my_label2}
\end{table*}

\section*{Sensitivity to Frame Rate}
The dataset used in this study contains a diverse sample of inputs with different frame rates and qualities. We aimed to design and train a network that would be robust to these types of variations. To evaluate this robustness against changes in frame rate, we created embeddings of several studies at original frame rate as well as their temporally down-sampled versions. In ~\autoref{fig:frsens}, a qualitative sample of the sensitivity of embedding quality to frame rate is visualized. In~\autoref{fig:FR_sync}, a screenshot of synced cines from AP4 and AP2 views with different frame rates are visualized. Please refer to the supplementary video for a more obvious correspondence of cardiac walls and chambers during a cardiac cycle with different frame rates.
This experiment shows the robustness of the proposed synchronization method to different frame rates.

\section*{Description of the Alignment Algorithm}
\begin{figure}
    \centering
    \includegraphics[width=\linewidth]{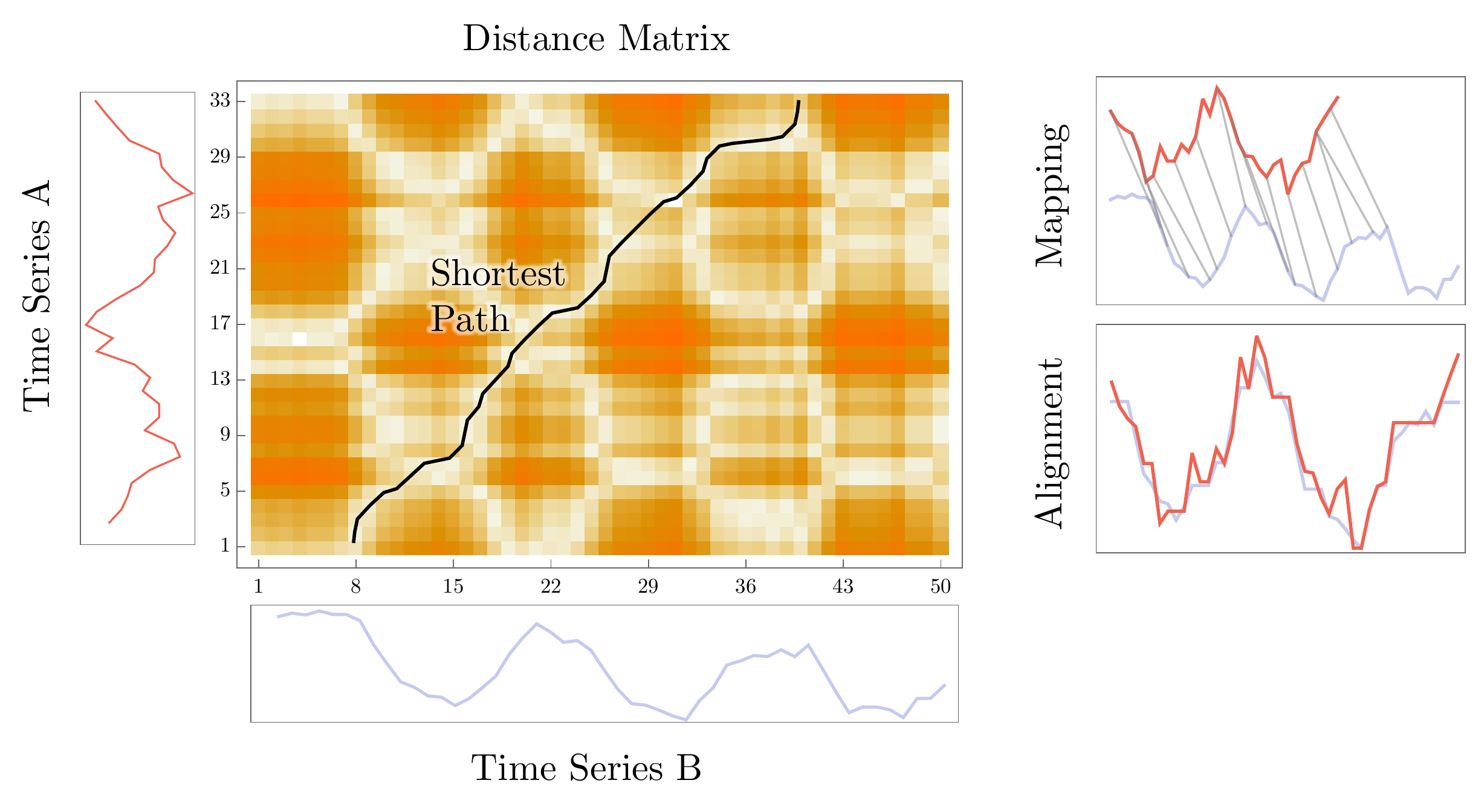}
    \caption{Visual description of the dynamic time warping alignment algorithm. The pairwise distance matrix is computed for time series A and B. The optimal alignment between the two series is framed as a shortest path problem across the smaller dimension of the distance matrix---this is shown by the black curve. Each point on the black curve represents the mapping between series A and B (top right). The alignment is found by gluing together elements that are mapped to each other (bottom right).}
    \label{fig:dtw}
\end{figure}

Echo-SyncNet produces 128-dimensional embeddings vectors for each frame of a given input cine. Two given cines can be temporally aligned by computing the dynamic time warping (DTW) \cite{dtw} correspondence between their embedding vectors. DTW is a well known dynamic programming algorithm for time series alignment. The algorithm computes the point-wise correspondence between two temporal sequences by minimizing the sum of the distance between every pair of corresponding points while preserving temporal ordering---a visual description of the algorithm is shown in \autoref{fig:dtw}. The complexity of the algorithm is $O(NM)$, where $N$ and $M$ are the lengths of the two sequences, respectively. To align more than two sequences together using DTW, it suffices to choose one as a reference and compute the warping correspondences between it and all targets. 

\bibliographystyle{IEEEtran}
\bibliography{main.bib}

\end{document}